\documentclass{article}
\usepackage{spconf,amsmath,epsfig}
\usepackage{pbox}
\usepackage{tikz}
\usepackage{pgfgantt}
\usepackage{subfigure}
\usepackage{gensymb}
\usepackage{amsmath}
\usepackage{adjustbox} 
\usepackage{amsfonts}

\usepackage{array}
\usepackage{graphicx}
\usepackage{multirow}

\usepackage{soul}

\usetikzlibrary{matrix,shapes,arrows,positioning,chains}
\tikzstyle{process} = [rectangle, minimum width=3cm, minimum height=1cm, text centered, draw=black]
\tikzstyle{arrow} = [thick,->,>=stealth]
\tikzstyle{io} = [trapezium, trapezium left angle=70, trapezium right angle=110, text centered]
\tikzstyle{decision} = [diamond, draw, text  centered, inner sep=1pt]

\title{TRUSTING SMALL TRAINING DATASET FOR SUPERVISED CHANGE DETECTION}
\name{Sudipan Saha$^{1}$, Biplab Banerjee$^{2}$, Xiao Xiang Zhu$^{1,3}$}

\address{Data Science in Earth Observation, Technical University of Munich, Taufkirchen/Ottobrunn, Germany$^{1}$ \\ Indian Institute of Technology Bombay,  Mumbai, India$^{2}$ \\ Remote Sensing Technology Institute, German Aerospace Center (DLR),  We{\ss}ling, Germany$^{3}$}

\begin{document}

%\ninept
%
\maketitle
\begin{abstract}
Deep learning (DL) based supervised change detection (CD) models require large labeled training data. Due to the difficulty of collecting labeled multi-temporal data, unsupervised  methods are preferred in the
CD literature. However, unsupervised methods cannot fully exploit the potentials of data-driven deep learning and thus they are not absolute alternative to the supervised methods. This motivates us to look deeper into the 
supervised DL methods and investigate how they can be adopted intelligently for CD by minimizing the requirement 
of labeled training data. Towards this, in this work we show that
geographically diverse training dataset can yield significant improvement over less diverse training datasets of the same size. We propose a simple confidence indicator for verifying the trustworthiness/confidence of supervised models trained with 
small labeled dataset. Moreover, we show that for the test cases where supervised CD model is found to be less confident/trustworthy, unsupervised methods often produce better result than the supervised ones.

\end{abstract}
%

%%Frames added to image - thickness
\fboxsep=0mm%padding thickness
\fboxrule=0.1pt%border thickness

\footnote{Accepted at IEEE Geoscience and Remote Sensing Symposium 2021}

\section{Introduction}
\label{secIntro}
Change detection (CD) is a topic of paramount importance in remote sensing. It is used in several applications, including disaster management \cite{saha2020unsupervisedSentinel2Grsl} , urban monitoring \cite{daudt2019multitask},  
agriculture \cite{saha2020unsupervisedSentinel2Grsl}, and forestry \cite{saha2020unsupervisedSentinel2Grsl}. Deep learning based change detection methods can be supervised \cite{mou2018learning}, unsupervised \cite{saha2019unsupervised} or semi-supervised \cite{saha2020semisupervised}.  Moreover, deep learning based CD methods have been proposed for both active  and passive sensors \cite{daudt2019multitask, mou2018learning, saha2019unsupervised}.
\par
 Supervised deep learning methods for CD depend on the availability of labeled multi-temporal training samples \cite{daudt2019multitask, mou2018learning, zhang2018change}. 
 It is inherently difficult to collect labeled data in the context of multi-temporal analysis \cite{saha2019unsupervised}. 
Due to the lack of training data, some of the supervised CD models are trained and tested on pixels from same image \cite{mou2018learning}.
  Hence, to alleviate the requirement of labeled training data, a number of unsupervised methods have been proposed in literature \cite{saha2019unsupervised, saha2020unsupervised}.
 They either reuse the deep models originally trained for image classification \cite{saha2019unsupervised, saha2020unsupervisedSentinel2Grsl} or train multi-temporal model using self-supervision \cite{saha2020unsupervised}. However, it obvious  that unsupervised CD methods are not as good as the supervised ones when 
 sufficient training data is available \cite{saha2020unsupervisedSentinel2Grsl, daudt2019multitask}. While unsupervised CD methods work well in most applications, they are not absolute alternative to the supervised methods. 
 \par
 While emphasizing the importance of supervised CD methods, we reiterate the difficulty to collect abundant labeled multi-temporal data for most CD applications. This difficulty is even 
 aggravated by the fact that remote  sensing operates across a significant variation of geography and deals with a large number of sensors. Under this circumstances, it is 
 of utmost importance to understand how to collect training data such that minimal training data can yield more effective supervised CD model. Active learning \cite{liu2016active} is often
 used to seek data-efficient training data by embedding data selection within the learning mechanism. Differently from active learning, our aim is to find a general guideline suitable for multi-temporal label
 collection without modifying the learning mechanism. We hypothesize that the geographic variation in training data may be instrumental to generate more robust supervised CD models.
 It is also crucial to understand 
 when a deep learning based CD model can be trusted. Towards this, we point towards the recent development in the literature related to predictive uncertainty \cite{malinin2018predictive}.
It has been shown that deep learning based models are prone to predictive uncertainties \cite{malinin2018predictive} from three different sources, e.g., model  
uncertainty, data uncertainty, and distributional uncertainty. CD models can be prone to all of them, model uncertainty due to the lack of significant training data, data
 uncertainty due to the overlap between changed and unchanged data, which may cause errors in labeling. Moreover, distributional uncertainty can arise in CD due to the geographic difference 
between training multi-temporal data and the test ones. Thus, while trained with very few training data, it is important to understand when we can trust the result obtained from the supervised deep
CD model. Towards this, we propose a simple index that can provide an indication of the trustworthiness of the supervised deep CD model. Under resource-constrained scenario where supervised CD models are
not trustworthy, we hypothesize that unsupervised CD models \cite{saha2020unsupervisedSentinel2Grsl, saha2019unsupervised} can be more reliable than the supervised ones. We show this to be true using a set of experiments.

\tikzset{
    between/.style args={#1 and #2}{
         at = ($(#1)!0.5!(#2)$)
    }
}
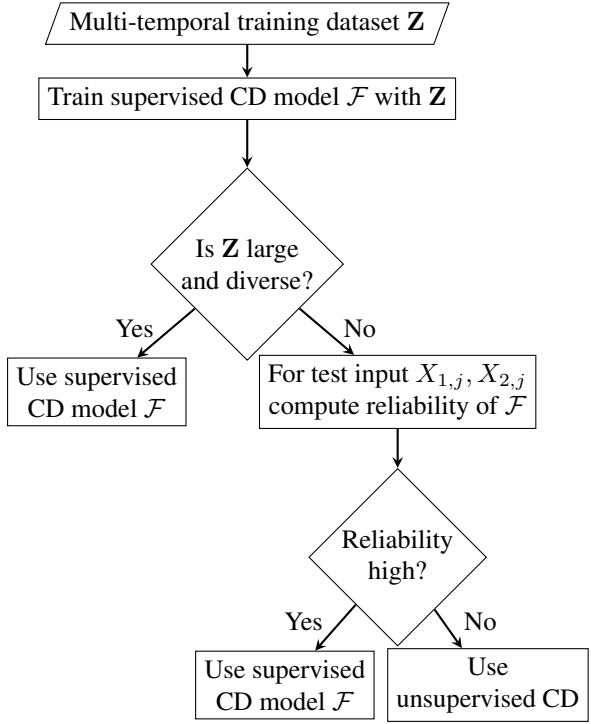
\begin{figure}[h]
 \centering
 \begin{tikzpicture}
 \node (1) [draw, io, align=center] { Multi-temporal training dataset $\mathbf{Z}$};
  \node (2) [draw, align=center,below of=1] {Train supervised CD model $\mathcal{F}$ with $\mathbf{Z}$};
\node (3) [decision,draw, align=center, below of=2,yshift=-1.20cm]{Is $\mathbf{Z}$ large\\ and diverse?}; 
\node (4a) [draw, align=center,  below of=3,yshift=-0.70cm,xshift=-2cm] {Use supervised\\ CD model $\mathcal{F}$};
\node (4b) [draw, align=center,  below of=3,yshift=-0.70cm,xshift=2cm] {For test input $X_{1,j}, X_{2,j}$\\compute reliability of $\mathcal{F}$};
\node (5b) [decision,draw, align=center, below of=4b,yshift=-1.20cm]{Reliability\\ high?}; 
\node (6ba) [draw, align=center,  below of=5b,xshift=-1.5cm,yshift=-0.70cm] {Use supervised\\ CD model $\mathcal{F}$};
\node (6bb) [draw, align=center,  below of=5b,xshift=1.2cm,yshift=-0.70cm] {Use \\ unsupervised CD};

\node (auxiliaryBetween3And4a) [between=3 and 4a,xshift=-.5cm]{Yes}; 
\node (auxiliaryBetween3And4b) [between=3 and 4b,xshift=.5cm]{No}; 

\node (auxiliaryBetween5bAnd6ba) [between=5b and 6ba,xshift=-.5cm]{Yes}; 
\node (auxiliaryBetween5bAnd6bb) [between=5b and 6bb,xshift=.5cm]{No};

\draw [arrow] (1) -- (2);
\draw [arrow] (2) -- (3);
\draw [arrow] (3) -- (4a);
\draw [arrow] (3) -- (4b);
\draw [arrow] (4b) -- (5b);
\draw [arrow] (5b) -- (6ba);
\draw [arrow] (5b) -- (6bb);

\end{tikzpicture}
 \caption{Proposed approach for deciding between using supervised and unsupervised CD method}
 \label{figureSupervisedOrUnsupervisedLearning}
\end{figure}

\section{Proposed approach}
\label{sectionProposedMethod}
Let us assume that we have a set of multi-temporal training scenes $\mathbf{Z} = \{Z_{1,i}, Z_{2,i}$\} $(i=1,...,I)$ where I is the size of the training dataset. The images in the training dataset is used to learn a 
supervised CD model $\mathcal{F}$. Once trained, $\mathcal{F}$ can be used to predict CD labels from the test scenes $X_{1,j}, X_{2,j}$ $(j=1,...,J)$. In CD, 
number of distinct labels are $K+1$:  $K$ labels corresponding to different kinds of change ($\omega_{c1}, \omega_{c2}, ..., \omega_{cK}$) and one corresponding 
to unchanged pixels $\omega_{nc}$ \cite{saha2019unsupervised}. For binary CD, this changed classes merge to one class - $\Omega_{c}$. This work is a preliminary work where we
discuss about how images in $\mathbf{Z}$ can be chosen to make $\mathcal{F}$ more robust. Moreover we propose an indicator which can give us a relative indication about trustworthiness of $\mathcal{F}$
on the test images. We postulate that when supervised model $\mathcal{F}$ is not trustworthy, unsupervised methods \cite{saha2019unsupervised} are more preferable than the supervised ones. Proposed scheme is shown in
Figure \ref{figureSupervisedOrUnsupervisedLearning}.

\textit{Supervised model architecture}: For training $\mathcal{F}$, any suitable architecture can be used. In this, we chose the fully convolutional network (FCN) as in \cite{daudt2019multitask}. FCN architecture ingests as input
the concatenation of the  pre-change and post-change images. Since there is no fully connected layer in the architecture, FCN architecture is able to ingest input of any spatial size.
FCN architecture is trained end-to-end. 

\textit{Geographic variation enables robust model}: Variation in the training data enables us to obtain more robust model. This is one of the reason that data augmentation techniques are popular in computer vision, in addition to the reason that they help in increasing the size of the dataset. In multi-temporal remote sensing, variation may come in different form, e.g., usage of different sensor and different geographic places. While the other variations may also be crucial to obtain
better models, here we solely focus on the geographic variation. Let us assume that we have two different instances of the training dataset: $\mathbf{Z}_1$  and $\mathbf{Z}_2$, both consisting of  $I$ images.
$\mathbf{Z}_1$ contains images that geographically diverse, i.e., taken from different cities/countries. On the other hand, $\mathbf{Z}_2$ contains images that are geographically much more localized, i.e., less diverse than
$\mathbf{Z}_1$. $\mathbf{Z}_1$ and $\mathbf{Z}_2$  are separately used to train models  $\mathcal{F}_1$ and $\mathcal{F}_2$, respectively. They are trained
using the same architecture and same training protocols. We hypothesize for a test pair  $X_{1,j}, X_{2,j}$ which is not geographically covered by
both $\mathbf{Z}_1$  and $\mathbf{Z}_2$, $\mathcal{F}_1$ will obtain superior result in comparison to $\mathcal{F}_2$.  The gap in performance of $\mathcal{F}_1$ and $\mathcal{F}_2$ narrows down as
the size of the datasets ($I$) increases.

\textit{Confidence indicator}: As discussed in the previous paragraph, obtaining geographically diverse dataset may enable us to obtain more
robust supervised CD model. However, in practice, this is not always possible. In such scenario, 
it is important to have an indicator from the model when not to trust its output. A new pool of work is emerging the machine learning regarding
predictive uncertainty \cite{malinin2018predictive}. Most works regarding predictive uncertainty require additional training via multi-tasking for
uncertainty estimation. Our goal is to obtain a reliability indicator without using any such explicit training. Towards this, we take inspiration from the literature
related to the uncertainty estimation, especially from  Dirichlet Prior Network (DPN) \cite{malinin2018predictive} where exponential of logits  of the deep
network (also called concentration parameters) are used to characterize the DPN.  Though we do not model our deep network as DPN, we follow similar strategy and use the logits as an
indicator of the relative confidence of the network. Logits are the outputs before applying softmax to the deep CD model $\mathcal{F}$. 
Since, we use FCN architecture, logit is obtained from
all pixels of the analyzed scene. For a pixel $x^{*}$ in consideration, for $K+1$ outputs in multiple CD,  we obtain
    logits as $\{z_0(x^{*}),z_1(x^{*}),...,z_K(x^{*})\}$. In case of binary CD, we obtain two logits as $\{z_0(x^{*}),z_1(x^{*})\}$. For a confident prediction, one of the two will be significantly larger than the other and hence
the maximum of the two is taken as $z_{opt}(x^{*})$. Confidence indicator ($\beta_j$) is obtained as mean of the $z_{opt}(x^{*})$ obtained from all pixels. Let us assume that we are 
interested to apply $\mathcal{F}$ on different test scenes: scene 1 comprising of bi-temporal images $X_{1,j1}, X_{2,j2}$ and scene 2 comprising of
bi-temporal images $X_{1,j2}, X_{2,j2}$. Let, $\beta_{j1}$ and $\beta_{j2}$ be the confidence indicator estimated from them. We hypothesize that result obtained by $\mathcal{F}$
is more trustworthy on the first scene than the second. Finally, relative confidence indicator $\beta'_j$ is obtained by min-max normalization of $\beta_j$ $(j=1,...,J)$.

\textit{Unsupervised CD as alternative}: Though unsupervised CD methods cannot be an absolute alternative to the supervised methods, we postulate that for the test cases
where supervised methods yield low confidence, unsupervised methods
can produce superior result. The indicator proposed above can indicate test cases where the trained network $\mathcal{F}$ is not
confident about its prediction. In such scenario, supervised methods can be replaced by the unsupervised methods, e.g., those reusing models trained for image classification without any training on multi-temporal data \cite{saha2019unsupervised} or those using self-supervised multi-temporal learning \cite{saha2020unsupervised}.

\section{Experimental Results}
\label{sectionExperimentalResult}
\textit{Dataset and settings:} To verify our propositions, we require a dataset where training scenes are completely disjoint
from the test ones, similar to \cite{daudt2018urban}, however unlike \cite{mou2018learning}. Furthermore, training scenes need to be distributed over different locations across the Earth. We choose the 
existing Sentinel-2 based Onera Satellite Change Detection (OSCD) dataset \cite{daudt2018urban} as it adheres to the
above properties. The dataset consists of 14 training scenes (image pairs) distributed across different countries.
As our objective is to understand the behavior of deep CD models when amount of labels are severely low, we restrict the number of training scenes to three. In our experiments, we considered
four 10 meter bands - R, G, B, NIR. Quantitative results are computed as sensitivity (accuracy computed over changed pixels), specificity
(accuracy computed over unchanged pixels), and kappa \cite{saha2019unsupervised}

\textit{Geographical variation:} To test the hypothesis that geographical
 variation enables robust model, we choose four different combinations of three cities for training -
 \begin{enumerate}
 \item \textit{Localized 1}: Bercy, Rennes, Saclay (E).
 \item \textit{Localized 2}: Bordeaux, Nantes, Paris.
 \item \textit{Diverse 1}: Nantes, Hongkong, Beirut.
 \item \textit{Diverse 2}: Beihai, Hongkong, Mumbai.
  \end{enumerate}
In the above combinations, the first two are geographically localized, as all training cities are sampled from small/big urban regions of France. On the other hand, the last two combinations 
consist of cities from different parts of the world and thus show diverse geographic condition. 
\par
\begin{figure}[!h]
\centering
\subfigure[]{%
            %\label{fig:first}
            \includegraphics[height=4 cm]{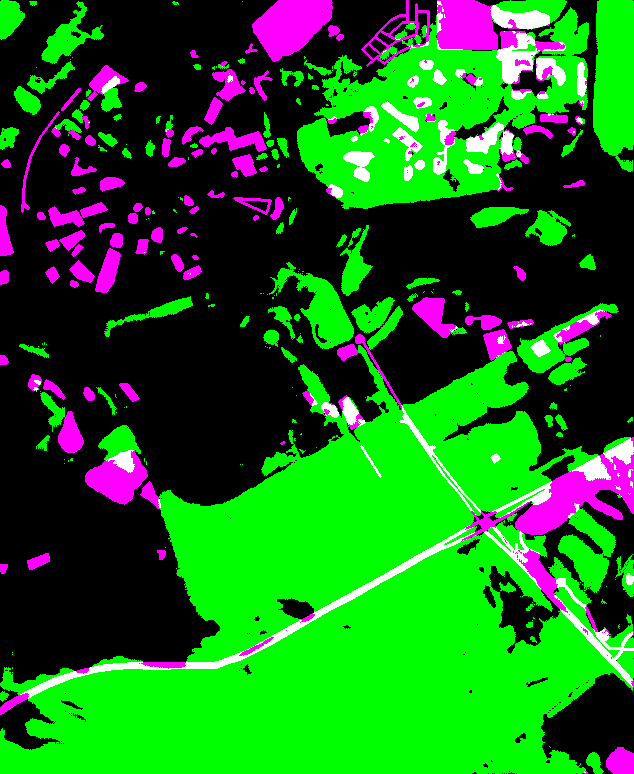}
            \label{inputImageWorldviewPre}
        }%
\subfigure[]{%
            %\label{fig:first}
            \includegraphics[height=4 cm]{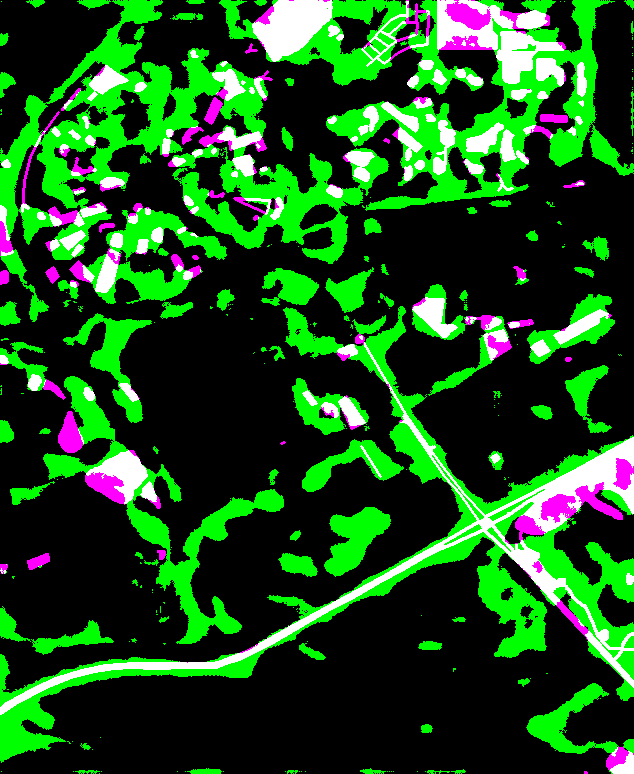}
            \label{inputImageWorldviewPost}
        }%SS

\caption{Visual result for Dubai when using training city combination: (a) ``localized 1'' and (b) ``distributed 1''. Maps are shown as false color composition where white color indicates correctly detected changed pixels.
Green and pink colors indicate false alarms and missed alarms, respectively. It is evident that the first case misses most of the changes and generates a significant amount of false alarm.
While in the latter case, a large number of changes are correctly detected.}
\label{figureVisuaDubai}
%%Be careful about this command added intentionally to save space
\end{figure}

We tested the models trained with above cities as training scenes on the test cities in OSCD dataset.  The quantitative result (Table \ref{tableGeographicVariationTrainingCities}) clearly shows that two
diverse combinations produce much better result than their localized counterparts. Surprisingly, diverse training dataset obtains superior result even for  test city ``Montpellier'', which is
geographically in same region as the cities in training set ``localized 1'' and ``localized 2''. This shows that diversity is useful not only for test sets which are markedly distinct from the training
set, but also for those which has strong resemblance to the training set.  We show the result for test city Dubai as false color composition between detected CD and reference CD map
in Figure \ref{figureVisuaDubai}.
 It is evident that ``distributed 1'' obtains superior performance to ``localized 1''.

\textit{Confidence indicator:} When training data is very few and localized, as in for ``localized 1'' case, it is important to get an indication from the supervised CD model when it is 
failing to provide good result. In Table \ref{tableReliabilityIndicator}, we show the CD accuracies and proposed confidence indicator for different cities. We clearly observe that the proposed reliability indicator can 
provide a rough indication of where the CD result can be trusted. Confidence indicator is much higher for cities that obtained higher kappa value compared to those that obtained lower kappa value.
Thus the proposed confidence indicator can be used for relative assessment of supervised CD model.

\textit{Unsupervised CD for low-confidence test cases:} In \cite{saha2020unsupervisedSentinel2Grsl}, deep unsupervised CD method obtains inferior result in comparison to the supervised methods (considering all test cities).
Aligned with this, for test city ``Montpellier'' method in  \cite{saha2020unsupervisedSentinel2Grsl} obtains a kappa score of 0.26, much lower than the supervised method (Table \ref{tableReliabilityIndicator}) with training set
``localized 1''. However, for the test city ``Dubai'', method in \cite{saha2020unsupervisedSentinel2Grsl} obtains a kappa score of 0.18, thus outperforming the supervised method (Table \ref{tableReliabilityIndicator}) where supervised method has low
confidence score.

\renewcommand{\tabcolsep}{2pt}
\begin{table}
\centering
\caption{Quantitative binary CD result with different combinations of training cities. Best Kappa for each cimbination is shown in bold.}
\begin{tabular}{|c|c|c|c|c|} 
 \hline
\textbf{Test city} & \textbf{Training cities} & \textbf{Sensitivity}  & \textbf{Specificity} & \textbf{Kappa}  \\ 
\hline
\bf \multirow{4}{*}{All} & Localized 1 & 54.03 & 78.98 & 0.12 \\ 
\cline{2-5}
& Localized 2 & 42.39 & 88.94 & 0.18  \\ 
\cline{2-5}
& Diverse 1 & 71.80 & 86.38 & 0.28  \\ 
\cline{2-5}
& Diverse 2 & 57.52 & 91.55 & \textbf{0.32} \\ 
\hline
\bf \multirow{4}{*}{Montpellier} & Localized 1 & 47.41 & 99.06 & 0.57 \\ 
\cline{2-5}
& Localized 2 & 33.83 & 98.85 & 0.43  \\ 
\cline{2-5}
& Diverse 1 &  66.87 & 96.70 & \textbf{0.60}  \\ 
\cline{2-5}
& Diverse 2 & 41.25 & 99.17 & 0.52  \\ 
\hline
\bf \multirow{4}{*}{Milano} & Localized 1 & 21.95 & 99.01 & 0.17 \\ 
\cline{2-5}
& Localized 2 & 58.78 & 97.98 & 0.28  \\ 
\cline{2-5}
& Diverse 1 & 76.89 & 97.12 & 0.28  \\ 
\cline{2-5}
& Diverse 2 & 44.53 & 99.03 & \textbf{0.33}  \\ 
\hline
\bf \multirow{4}{*}{Rio} & Localized 1 & 31.53 & 98.30 & \textbf{0.37} \\ 
\cline{2-5}
& Localized 2 & 18.52 & 98.18 & 0.22  \\ 
\cline{2-5}
& Diverse 1 & 40.35 & 96.19 & 0.36  \\ 
\cline{2-5}
& Diverse 2 & 27.68 & 98.22 & 0.32  \\ 
\hline
  \end{tabular}
\label{tableGeographicVariationTrainingCities}
\vspace{-0.5cm}  %%Be careful about this command added intentionally to save space
\end{table}

\renewcommand{\tabcolsep}{2pt}
\begin{table}
\centering
\caption{Relationship of CD performance (shown as kappa) and the reliability indicator $\beta'_j$ for training combination ``localized-1''}
\begin{tabular}{|c|c|c|c|c|} 
 \hline
\textbf{Test city} & \textbf{Sensitivity}  & \textbf{Specificity} & \textbf{Kappa} & \textbf{Reliability ($\beta'_j$)}  \\ 
\hline
Montpellier & 47.41 & 99.06 & 0.57 & 1  \\   %%1.955
\hline
Rio & 31.53 & 98.30 & 0.37 & 0.95  \\   %%1.949
\hline
Brasilia & 15.55 & 97.63 & 0.13 & 0.58  \\   %%1.90
\hline
Valencia & 20.95 & 96.20 & 0.04 & 0.45  \\   %%1.883
\hline
Dubai & 7.12 & 96.81 & 0.06 & 0  \\  %%1.823
\hline
  \end{tabular}
\label{tableReliabilityIndicator}
\vspace{-0.15cm}  %%Be careful about this command added intentionally to save space
\end{table}

\section{Conclusions}
\label{sectionConclusion}
Our work is an attempt towards understanding the reliability of multi-temporal labels for supervised CD. Towards this, we showed that geographically distributed training dataset can obtain superior result compared
to the geographically confined ones, given the
application is fixed. We also proposed a simple yet effective indicator which denotes the confidence of the deep supervised CD model about its prediction. We further showed that when deep supervised CD shows low confidence, unsupervised
methods can outperform the supervised ones.  It is very critical to design self-aware AI systems that intelligently use training labels and be aware of their prediction's uncertainty. Our work
 is a preliminary step towards that in context of deep supervised CD.
In future, we would like to take this further by analyzing the impact of geographic variation in more details.  We would like to design a normalized and more robust confidence indicator for deep multi-temporal models. Furthermore, we would like to experiment on different architectures.

\section*{Acknowledgement}
\label{sectionAcknowledgement}
The work is funded by the German Federal Ministry of Education and Research (BMBF) in the framework of the international future AI lab ``AI4EO -- Artificial Intelligence for Earth Observation: Reasoning, Uncertainties, Ethics and Beyond" (Grant number: 01DD20001).

%\bibliographystyle{IEEEbib}
\bibliographystyle{ieeetr}
\bibliography{sigproc}

\end{document}